# SIGNSWORLD; DEEPING INTO THE SILENCE WORLD AND HEARING ITS SIGNS (STATE OF THE ART)

[1]A.M. Riad, [2]Hamdy K.Elmonier, [1]Samaa.M.Shohieb, and [1]A.S. Asem

[1]Faculty of Computer and Information Sciences.Information Systems department
[1]Sm.shohieb@yahoo.com

[2]Misr academy for Engineering and Technology
[2]hamdy_elminir2@yahoo.com

*ABSTRACT*.

*Automatic speech processing systems are employed more and more often in real environments. Although the underlying speech technology is mostly language independent, differences between languages with respect to their structure and grammar have substantial effect on the recognition systems performance. In this paper, we present a review of the latest developments in the sign language recognition research in general and in the Arabic sign language (ArSL) in specific. This paper also presents a general framework for improving the deaf community communication with the hearing people that is called SignsWorld. The overall goal of the SignsWorld project is to develop a vision-based technology for recognizing and translating continuous Arabic sign language ArSL.*

*KEYWORDS*

*Arabic Speech Recognition, Deaf-To-Normal Communication, Sign Language Recognition, Normal to Deaf Communication*

## 1. INTRODUCTION

There has been a surging interest in recognizing human hand gestures lately. Sign language is the most structured set of gestures and it is the primary means of communication among hard of hearing people.

The strong rule of context and grammars make sign language powerful enough to fulfil the needs of the deaf people in their day to day life. Sign language (SL) is a subset of gestural communication used in deaf-muted community, in which postures and gestures have assigned meanings with a proper grammar. Like any other verbal language, its discourse comprises of well structured rendering and reception of non-verbal signals according to the context rules of the complex grammar. Postures are the basic units of a sign language, and when collected together over a time axis and arranged according to the grammar rules, they reflect a concrete meaning [1].

To make a successful communication between the deaf and normal people, there must be two ways of communication; from deaf-to-normal and vice versa. In this paper we tried to cover the researches that were conducted in each way of communication.

From deaf-to-normal communication needs gestures, facial expressions and body language recognition techniques. These techniques are divided in two main types device based [2, 3] approaches and vision based approaches [16, 31, 34, 36, 45]. Each is also divided in its turn to manual signs; that include the hand and arms signal; recognition techniques [7, 37 ,43, 44] and non-manual signs; that include facial expressions and body language; recognition techniques[1,46,47].

Otherwise, from normal-to-deaf communication needs speech [57, 12, 64] and visual speech [81,85,86] recognition techniques. Also for a good communication there is a need to include the emotion recognition techniques [32, 37,38].

The second section of this paper presents the deaf to normal communication techniques that have been presented by another research studies. The third section presents the normal to deaf communication techniques. In the fourth section authors presents the proposed SignsWorld project specification, architectures and challenges. Finally; the conclusions and the future work will be presented.

## 2. DEAF-TO-NORMAL COMMUNICATION

According to Rung-Hui et al. [4] a posture is a static gesticulation of an articulator (hand, eyes, lips, body) while a gesture is a sequence of postures having defined meaning in a particular SL. For example, finger spelling in any sign language is communicated by making postures (static signs) or each letter ("a", "b", "c" etc) but in the case of continuous discourse, most signs comprise of gestures (dynamic signs). Apart from the temporal classification of signs, they can also be categorized according to their major articulators. Manual signs are performed using hands while non-manual signs (NMS) mainly include facial expressions, body movement and torso orientation. Although manual signs constitute a large proportion of sign language vocabulary, NMS also own a significant share to convey the whole context. Hence it differs from a spoken language in a way that a spoken language structure uses the words in a sequential manner but the SL structure allows manual and non-manual components to be performed in parallel. Another unique feature of SL over any spoken language is its capability to convey multiple ideas at a single instant of time [1, 5].

### 2.1. Device-Based Approaches

All detection and recognition is performed on a set of data incoming to a processor unit from multiple sensor streams. All information related to signing articulators is captured by sensors or trackers. Sensors are worn by the signer and they may include displacement sensors, positional sensors or trackers. When a signer performs signing, articulator's data is captured on a specific rate and fed to the recognition stage. Unlike vision based methods, these schemes are efficient and robust due to smaller vocabulary set [55]. On the other hand, they severely affect the user independence due to a dense mesh of installed sensors [1, 5].

#### 2.1.1. Continuous Manual Signs Recognition

Manual signs (MS) are those visual signals which are not conducted by hand or arms A DataGlove [6] with multiple electronic sensors (installed on the finger joints, wrist and palm) is shown in Fig. 1 , which feeds these measurements in real time to a processing unit [7][8] The processing unit compares the set of static sign samples with existing templates and generates output.

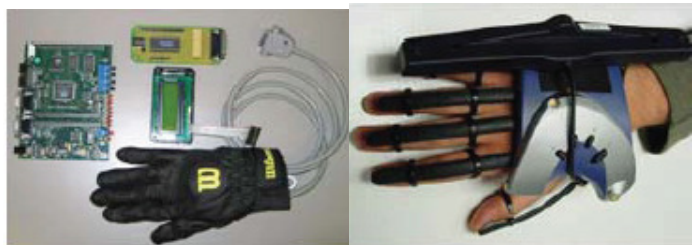

Figure 1.    DataGloves [9][10]

Static signs are easy to incorporate due to their defined boundaries but in the case of continuous signing, a 2D motion trajectory is formed and matched with an existing template or a learnt model. Most of these researches involve recognition of a small set of words, mainly static

190

postures or finger spellings. In another approach, a glove fitted with 7sensors is used for sign detection [2]. Out of 7 sensors, 5 were dedicated for finger joints and 2 for tilting and rotation of hand. Each sensor returns a discrete value from 0 to 4095, where 0 indicates fully open and 4095 for the fully bent state. The sample rate was 4 Hz and the minimum sign hold duration was 750 mSec.

Braffort proposed a French Sign Language recognition system using HMMs [11]. A data glove was used to obtain hand appearance and position. Features of hands are extracted from their appearance and position. Signs are separated into conventional signs, non conventional signs, and variable signs. The system achieved a recognition rate of 96 percent with a vocabulary of seven signs. The experiments were performed with a continuous gesture data set.

Gao et al. proposed a Chinese Sign Language recognition system, which was based on a dynamic programming method [3]. The system used two data gloves and three position trackers to extract the hand appearance and position. The DTW method was used to match an input data and templates of movement epentheses. The system could recognize 5,113 signs and achieved a recognition rate of 90.8 percent with 1,500 sentences. The experiments were performed with a continuous data set. It was assumed that movement epentheses between two signs are always similar in different sentences. However, movement epentheses between two signs vary in real-world applications.

Vogler and Metaxas described an American Sign Language (ASL) recognition system based on HMMs [12]. Three video cameras were used. An electromagnetic tracking system was used to extract 3D parameters of the signer's arm and hand. Two experiments were performed, both with 99 test sentences and a vocabulary of 22 signs. The system achieved a recognition rate of 94.5 percent for single signs. It achieved a recognition rate of 84.5 percent for complete sentences. The experiments were performed with a continuous data set. Wearable sensor based methods are suitable for smaller vocabulary recognition systems for static signs but they incurs an unavoidable burden of system's weight and interfaces which may affect the natural way of signing.

Van R. Culver in [13] described an isolated sign language recognition (SLR) system that combines features from a video camera and an instrumented glove. These features were tested on American Sign Language (ASL) vocabularies ranging from 10 to 200 Words and proved an accuracy about 94%. But it also is not practical in the real world.

Perrin [14] proposed a low cost and short vocabulary single laser-based gesture recognition system in which a laser beam is transmitted and the amount of reflected energy is correlated with a reference signal to measure the displacement from the point of contact. Electronically controlled micro mirrors are installed to direct the maximum amount of reflected energy towards the receiver. In other words, these mirrors align themselves in the direction of signing hand. The current hand position is extrapolated using the velocity information and the current state of the micro mirrors from the previous known states. In continuous signing, the 2D hand position is estimated with the help of the mirror's direction while the depth information is calculated through the approximation of displacement. For recognition, these positional parameters are matched with existing templates or fed to other classifiers. This scheme doesn't need any sensor to be worn and is suitable exclusively for those manual signs whose meanings are mainly reflected by hand or arm motion and velocity, not by finger flexion or orientation.

### 2.1.2 Non-Manual Signs Recognition

Non manual signs (NMS) are those visual signals which are not conducted by hand or arms but they are shown by facial expressions and body movements. Embedded sensor based gloves (DataGloves) are quite efficient for the detection and recognition of static signs of a short vocabulary language but they severely affect the natural way of signing by imposing the cumbersome electronics to be worn. Moreover these arrangements are not suitable to incorporate the non-manual movements of facial articulators (eyes, eye-brows etc). The same is true for laser based tracking methods [1].

## 2.2. Vision-Based Approaches

Vision-based sign language recognition approaches need hand detection and tracking algorithms to extract hand locations. Color, motion, or edge information is generally used to detect hands from input data. Vision-based approaches have limitations according to imaging conditions such as background, illumination, clothing, and so on [15].

### 2.2.1 Continuous Manual Signs Recognition

Liwicki and Everingham [16] investigate the problem of recognizing words from video, fingerspelled using the British Sign Language (BSL) fingerspelling alphabet.They achieved about 98.5% of accuracy but they only recognized the continuous finger spelling in the words not the sign language.

In [17] Wang et.al Proposed A multilayer architecture of sign language recognition for the purpose of speeding up the search process. The generation of confusion set is realized by DTW/ISODATA algorithm. And they achieved accuracy of about 94%. But they only recognized the hand gestures not also the facial expression.

Agris et al. [18] describe a vision-based recognition system that quickly adapts to unknown signers. A combination of Maximum Likelihood Linear Regression and Maximum A Posteriori estimation was implemented and modified to consider the specifics of sign languages, such as one-handed signs with recognition accuracy of only 78.6% and they recognized the isolated signs not the continuous ones

Hieu and Nitsuwat in [19] presented an image preprocessing and revised feature extraction methods for sign language recognition (SLR) based on Hidden Markov Models (HMMs). Multi-layer Neural Network is used for building an approximate skin model by using Cb and Cr color components of sample pixels. Gesture videos are spitted into image sequences and converted into YCbCr color space and proved about 94% accuracy. This technique can't be used for the facial expressions and the non-manual signs.

Starner et al. presented an ASL recognition system based on HMMs [20]. The experiments involved two systems using 40 signs. The first system observed the signer using a camera mounted on a desk and achieved 92 percent accuracy. The second system observed the signer using a camera mounted on a cap worn by the signer and achieved 98 percent accuracy. Their experiments were performed with a continuous data set. Only hand motion features were used; hand shapes to disambiguate signs with similar hand motions were not used.

Bauer and Kraiss proposed a German Sign Language recognition system based on HMMs in which the signer wore simple colored gloves to obtain data [21]. Subunit HMMs were used to model signs. Two experiments were performed. In the first experiment, a 92.5 percent recognition rate was achieved for 100 signs. In the second experiment, subunit HMMs used in the first experiment also performed the sign language spotting task, without retraining for 50 new signs. A recognition rate of 81.0 percent was achieved with an isolated data set.

Holden et al. proposed an Australian Sign Language recognition system based on HMMs [22]. The system extracted the angle between two hands with respect to thehead, their directions of movement, roundedness, and size ratio. A 97 percent recognition rate at the sentence level and 99 percent at the word level were achieved.

Guilin Yao presented a method using One-Pass (OP) pre-searching before Viterbi recognition. The experiments are processed in the large vocabulary database. [23] The experiments were performed with a continuous data set. But the accuracy was reduced .Bowden et al. developed a British Sign Languagerecognition system based on Markov chains combined with independent component analysis [24]. The system extracted a feature set describing the location, motion, and shape of hands. A recognition rate of 97 percent was achieved with 43 signs.

M. Yang et al. proposed an ASL recognition system based on a time-delay neural network [25]. The system used motion information to extract hand positions. The recognition rate was 96.2 percent with 40 signs. The experiments were performed with an isolated data set.

Yang and Sarkar proposed an ASL spotting method based on CRFs [26]. The system used motion information as features and the Kanade-Lucas-Tomasi method to track the motion of salient corner points. The system extracted key frames from sentences in the training data. Then, the system labeled each key frame with a coarticulation pattern or sign pattern. The spotting rate was 85 percent with 39 signs, articulated in 25 different sentences.

Yang et al. proposed an ASL recognition method based on an enhanced Level Building algorithm [27]. An enhanced classical Level Building algorithm was used, based on a dynamic programming approach to spot signs without explicit coarticulation models. The recognition rate was 83 percent with 39 signs, articulated in 25 different sentences. This research only considered hand motion. It did not consider hand shape for recognizing signs.

Nayak et al. proposed an ASL recognition method based on a continuous state space model [28]. They used an unsupervised approach to extract and learn models for continuous basic units of signs, which are called signemes, from continuous sentences. They automatically extracted a signeme model given a set of sentences with one common sign.

Tsai and Huang [29] presented a vision-based continuous sign language recognition system to interpret the Taiwanese Sign
Language (TSL).They proved an accuracy of 92.6.

Theodorakis et al. [30] proposed a Product-HMMs (PHMM) technique to recognize the Greek Sign LanguageIsolated sign recognition rate increased by 8,3% over movement only models and by 1,5% over movement-shape models using multistream HMMs. Also Maebatake et al.[31] proposed a sign language recognition method by using a multi-stream HMM technique to show the importance of position and movement information for the sign language recognition. They conducted recognition experiments using 21,960 sign language word data. As a result, 75.6 % recognition accuracy was obtained with the appropriate weight (position: movement=0.2:0.8), while 70.6 % was obtained with the same weight.

Akmeliawati et al. [32] proposed a short vocabulary SL interpreting system which recognises manual signs. Hands are detected using their natural skin colours along with various other features (position, velocity etc) and matched with an existing set of templates. Other important methods of recognition include hidden Markov model (HMM) [33] [34] [35] and artificial neural network (ANN) classifiers [36]. Alvi et al [37] use a stochastic language model for sentence formulation by rearranging individually recognized static signs. Davis and Shah [38] use Finite State machine to recognize a smaller set of dynamic gestures. Skin color detection systems are severely affected by varying illumination, complex background, the signer's ethnicity(skin color), and articulator occlusion. To eliminate the lighting problem, instead of RGB, fixed threshold ranges for Cb and Cr (YCbCr space) were utilized in a controlled environment [39].

This system may seem suitable for laboratory experimentation due to tightly controlled lighting but it will result in a large error rate if installed at public places like hospitals, courts and shops. To avoid the false positives in skin colour recognition, colour coded gloves were introduced. There were different colours on different parts of gesticulated hand (palm, fingers and back) [40] so that each part can be identified independently. These sorts of schemes, although they restrict signer's independence, are more robust as compared to skin colour based methods.

Latest technological development has resulted in the advent of the state-of-the-art Time of Flight (ToF) cameras [101] which can acquire not only intensity image but also a depth image of the scene with precision of a few millimeters. Currently a few ToF camera based solutions are proposed which accurately spot the movement and orientation of articulators in 3D [41,42]. Although a ToF camera can produce intensity image, it normally has low lateral resolution which is unsuitable for processing fine details.

- **Motion Epenthesis**

International Journal of Computer Science & Information Technology (IJCSIT) Vol 4, No 1, Feb 2012

Few researchers have addressed the problem of non-sign patterns (which include out-of-vocabulary signs, epentheses, and other movements that do not correspond to signs) for sign language spotting [43], [20], [26].

This is because it is difficult to model nonsign patterns. As mentioned previously, Gao et al. assumed that movements between two signs are always similar in different sentences [43]. Ruiduo Yang et.al. [44] presented an approach based on version of a dynamic programming framework, called Level Building,to simulataneously segment and match signs to continuous sign language sentences in the presence of movement epenthesis (me).The accuracy of the proposed algorithm is about 83%.

Likewise, Yuan et.al [45] and Gao et.al [43] also explicitly modeled movement epenthesis and performed matching with both the sign and movement epenthesis models. The difference was that they adopted an automatic approach to cluster movement epenthesis frames in the training data first. Other than this, Yang and Sarkar [26] use conditional random fields to segment sentence by removing me segments, but their approach do not produce a recognition result. While explicit modeling of movement epenthesis have been shown to yield improved results, concerns of scalability of that approach remain; the number of possible me's is quadratically related to the number of signs in the dataset.

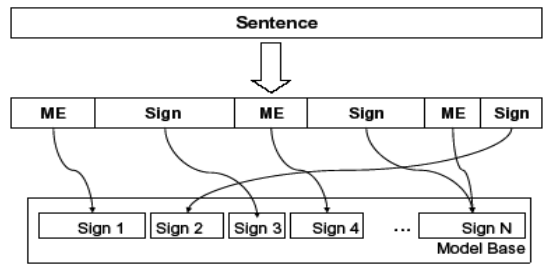

(a)

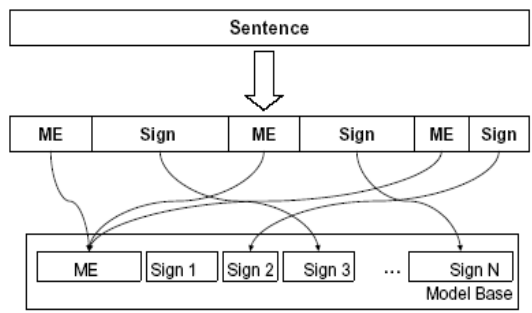

(b)

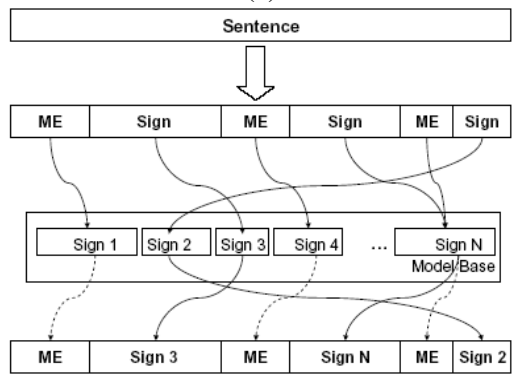

(c)

Figure. 2. Different approaches to handling movement epenthesis (me) in sentences: (a) If the effect of me is ignored while modeling, this will result in some me frames to be falsely classified as signs (b) If me is explicitely modeled, building such models will be difficult when the vocabulary grows large, since phonemes models are not well established for sign languages. (c) The adopted approach do not explicitly model mes instead, we allow for the possibility for me to exist when no good matching can be found. [44]

Yang and Sarkar extracted key frames for training coarticulatory movements from sentences consisting of in vocabulary signs and non-sign patterns [26].

Yang et.al. [55] proposed a method for designing threshold models in a conditional random field (CRF) model is proposed which performs an adaptive threshold for distinguishing between signs in a vocabulary and non-sign patterns. Their system can also achieve a 15.0 percent sign error rate (SER) from continuous data and a 6.4 percent SER from isolated data versus 76.2 percent and 14.5 percent, respectively, for conventional CRFs.

### *2.2.2. Non-Manual Signs Recognition*

Soontranon and Aramvith develop face and hand tracking for sign language recognition system. [46] The system is divided into two stages; the initial and tracking stages. In initial stage, they used the skin feature to localize face and hands of signer. The ellipse model on CbCr space is constructed and used to detect skin color. After the skin regions have been segmented, face and hand blobs are defined by using size and facial feature with the assumption that the movement of face is less than that of hands in this signing scenario. Their proposed method is able to decrease the prediction error up to 96.87% but increased in computational complexity up to 4%.

George Awad et al. [47] presented a unified system for segmentation and tracking of face and hands in a sign language recognition using a single camera combining 3 useful features: colour, motion and position. They used Kalman filter based algorithm and its error percentage was about 6.5%.

Vogler [48] presented a 3D deformable model tracking system to address this problem. They apply it to sequences of native signers, taken from the National Center of Sign Language and Gesture Resources (NCSLGR), with a special emphasis on outlier rejection methods to handle occlusions.

### *2.2.3. For the Arabic Sign Language*

To the best of our knowledge; there have been only three research studies who dealt with the Arabic finger spelling recognition [49][50][51]. And only one study. [52] dealt with the ArSL recognition problem.

In [49] developed a system to recognize the fingerspelling for the 28 alphabets using a colored gloves for data collection and an adaptive neuron-fuzzy interface system for recognition. A recognition rate of 88% was reached. In [50] M. Alrousan ,M. Hussain Alrousan used spatial features and a single camera to recognize the 28 alphabets without using gloves and reached to 93.55%. Also K.Assaleh and M. Alrousan [51] increased the recognition rate to 93.41% using polynomial networks.

Mohandes et.al. [52] propose an image based system for Arabic Sign Language recognition. A Gaussian skin color model is used to detect the signer's face. The recognition stage is performed using a Hidden Markov Model. The proposed system achieved a recognition accuracy of about 93% for a data set of 300 signs with leave one out method.

## 2.3. Symantec and Grammars Models

In a SL, continuous signs are detected and re-arranged in an appropriate order for a discreet context. Symantec and grammar models are very important referential models for complete syntax and semantics of a formulated sentence.

There is a lack in this point of research. In [53] a computational model of sign language utterances based on the construction of the signing space and a grammatical description: the iconicity theory.

Also SK Liddell in [54] studies of various aspects of the grammar of ASL left no doubt that signers using ASL were using a real human language.

Consequently; the requirement of a powerful grammar model ;especially in the Arabic language; is important. However, it will result in a computationally expensive solution.

## 3. NORMAL-TO-DEAF COMMUNICATION

The normal to deaf communication matter needs a speech recognition system with a noisy extraction and speaker's emotions extraction. For integrity also a visual speech recognition (lip reading) may also be needed. All of these points are discussed in the following subsections and ending them with the available Arabic speech recognition contributions.

### 3.1. Automatic Speech Recognition

Automatic Speech Recognition (ASR) [56] is a technique which machine can understand human's language. 2007 Science Publications The speech wave itself contains linguistic information that includes the meaning the speaker wishes to impart, the speaker's vocal characteristics and the speaker's emotion. Speech recognition is the process of automatically extracting and determining linguistic information conveyed by a speech wave using computers or electronic circuits. Only the linguistic information is needed from the speech wave, while the rest of the information is used in other fields of signal processing. A speech recognition system performs three primary tasks. Preprocessing: converts the spoken input into a form the recognizer can process. Recognition: identifies what has been said by comparing the input with the built-in reference models. Communication: sends the recognized input to the software systems that needs it.[57]

Taha et.al [58] proposed an agent-based design for Arabic speech recognition. They used a number of agents in the different recognition steps. The back propagation neural network achieved an average test accuracy of over (80.95) %.

Hesham Tolba et al [59] motivated the use of syllables to enhance the performance of automatic speech recognition (ASR) systems when dealing with large-vocabulary speech. Arabic and English are considered to test the proposed Approach using The HiddenMarkov Model Toolkit (HTK). the recognition rate using syllables outperforms the recognition rate using mono-phones and triphones by 40.08% and 19.74%, respectively.

Yong et al. [60] presented a method based expression recognition of audio-videos recognized by Fuzzy Buried Markov Model classifier. The average recognition rate of audio-visual expression is about 80.9%

In [61] Sah et al. compared three methods for speech temporal normalization namely the linear, extended linear and zero padded normalizations on isolated speech using different sets of learning parameters on multi layer perceptron neural network with adaptive learning. Although, previous work shows that linear normalization able to give high accuracy up to 95% on similar problem, the result in their experiment showed the opposite. They showed that The highest recognition rate using zero padded normalization is 99% while linear and extended linear normalizations give only 74% and 76% respectively.

Pan et.al. [62] presented an approach that can use few labeled Tibetan speech utterances to construct the effective recognition model. Their algorithm reached 88.6% accuracy, but the Vote Entropy of QBC and semi-supervised learning reached 84.8% and 82.5% respectively.

Patel and Rao [63] presented an approach to the recognition of speech signal using frequency spectral information with mel-frequency for the improvement of speech feature representation in a HMM based recognition approach. it is observed to have better efficiency for accurate qualification & recognition compared to the existing method.

*3.1.1 Speech Recognition in a Noisy Environment*

Speech is the natural way of communication so it should also be the way of communication with machines. However, and despite the high demand for speech recognition algorithms as biometric identifying tools, many speech recognition algorithms still have both False Acceptance Ratio (FAR) and high False Rejection Ratio (FRR) [64]. This particularly is the case when such systems are working in the real noisy environments. How to recognize and distinguish the voice from the environment and how to properly recognize the words from low quality voice samples are the main problems with which some of speech recognition algorithms and studies have been dealt.

Jie et al. [65] introduced an adaptive enhancement approach to resist the noises of different environments using Bark wavelet technique. They improved the SNR by about 9%.

In [64] Saeed and Szczepański worked with speech samples of nonhomogeneous quality. Their studies showed that cutting frequencies above 2200 Hz have rather low influence on the proper recognition but may rather lead to increase the errorrate.

Manikandan et al. [66] proposed an approach for phoneme classification using binary feature vector and correlation based classifier the proposed approach is simpler and requires lesser computational resources when compared with other pattern classification techniques.

Burileanu et.al. presents studies and early results with the scope to build a robust spontaneous speech recognition system in Romanian language. The final purpose of these attempts is to obtain substantial results in speech recognition for Romanian language that can be used as baseline for further results. [67]

Nolazo et al. [68] presented the Magnitude-Spectrum Information-Entropy (MSIE) of the speech signal. In two groups of noises the overall improvement performance in the range 0dB to 20 dB for the Aurora 2 database is of 15.06%.

Lishuang and Zhiyan [69] proposed a Chinese speech recognition system using integrating feature and Hidden Markov Model (HMM aiming at improving speech recognition rate in noise environmental conditions. The system is comprised of three main sections, a pre-processing section, a feature extracting section and a HMM processing section. Six Chinese vowels were taken as the experimental data. Using this technique decreased the Signal to Noise Ratio (SNR) by about 12%.

## 3.2. Emotions Recognition

Emotional speech recognition aims at automatically identifying the emotional or physical state of a human being from his or her voice. The emotional and physical states of a speaker are known as emotional aspects of speech [70] and are included in the so called paralinguistic aspects. Although the emotional state does not alter the linguistic content, it is an important factor in human communication, because it provides feedback information in many applications as it is outlined next. Making a machine to recognize emotions from speech is not a new idea. The first investigations were conducted around the mid-eighties using statistical properties of certain acoustic features [71].

Sebe et. al. described the problem of bimodal emotion recognition [72] and advocates the use of probabilistic graphical models when fusing the different modalities.they tested their audio-visual emotion recognition approach on 38 subjects with 11 HCI-related affect states. where the authors report average bimodal results of 89%.

Wenjing et. al. [73] constructed a VQ/ANN (vector quantization/artificial neural network) based speech emotion recognition system. The system first extracts the basic prosodic parameters and Mel-frequency cepstral coefficients (MFCC) frame by frame. Recent researches reveal that MFCC convey detailed emotional relevant information of syllable. However, the statistic measures of MFCC confuse the information at sentence level. They proposed a VQ-based method different to statistic method to generate measures of MFCC.the accuracy for Anger, Happiness, Sad, and Surprise are 69.5, 60.1, 94.2, 63.0 , and 71.7 respectively. Using the combination features, the recognition rate is 9.3% higher than by using the statistic features totally.

[74] Zhou et.al presented an approach that using articulatory features (AFs) derived from spectral features for speech emotion recognition.They investigated the combination of AFs and spectral features. Systems based on AFs only and combined spectral-articulatory features are tested on the,CASIA Mandarin emotional corpus.The recognition rate for happy neutral sad surprise are 64.00%,40.50%, 63.00%, 62.00%, and 47.50% respectively.

Chen et al [75] presented an audio visual multi-stream DBN model (Asy_DBN) for emotion recognition with constraint asynchrony, in which audio state and visual state transit individually in their corresponding stream but the transition is constrained by the allowed maximum audio visual asynchrony. angry happy neutral sad The recognition rate for angry, happy, neutral,and sad are 50.00%, 80.00%, 90.00% and 56.67 respectively for FullAsy_DBN approach and are 73.33%, 93.33%,, 96.67%,and 73.33 respectively for the Asy_DBN approach.

Qin et al. [76] proposed an emotional speech model based on the fuzzy emotion hypercube in order to support basic emotions in speech That varies from one application to another depending on their according needs. All emotions are recognised with an accuracy higher than 70%.

In [77] Huang et al adopted the missing feature theory and specified the Unreliable Region (UR) as the parts of the utterance with high emotion induced pitch variation. To model these regions, a virtual HD (High Different from neutral, with large pitch offset) model for each target speaker was built from the virtual speech. The recognition rate for BS , PDDM SS ,PDDM , PS and BSR are 51.55, 51.79, 52.59, and 53.43 respectively.

Tawari and Trivedi [78] presented a framework with adaptive noise cancellation as front end to speech emotion recognizer. They introduce feature set based on cepstral analysis of pitch and energy contours. Confusion using clean speech for EMODB database with 84.01% overall recognition rate Confusion using clean speech for LISA-AVDB database with 88.1% overall recognition rate

Wang et.al. [79] discussed Five common emotions, namely happiness, anger, boredom, fear and sadness, and recognized through a proposed framework which combines Principal Component Analysis and Back Propagation neutral network. The candidate parameters were refined from 43 to 11 via PCA to stand for a certain emotional type. Two neural network models, One Class One Network and All Class One Network, were employed and compared. The result, ranging from 52%-62%, suggests that the framework is feasible to be used for recognizing emotions in spoken utterance.

Xin and Xiang [80] proposed ECC via feature extraction of the Hilbert energy spectrum which describes the distribution of the instantaneous energy. The experimental results conspicuously demonstrated that ECC outperforms the traditional short-term average energy. Afterwards, further improvements of ECC were developed. TECC is gained by combining ECC with the Teager energy operator, and EFCC is obtained by introduced the instantaneous frequency to the energy. Seven status of emotion were selected to be recognized and the highest 83.57% recognition rate was achieved within the classification accuracy of boredom reached up to 100%.

### 3.3. Visual Speech Recognition

New HCI techniques emphasize on intelligent systems that can communicate with the users in a natural and flexible manner. The conventional human computer interfaces (HCI) such as mice and keyboards may not be suitable for people with limb disabilities. Users suffering from diseases or accidents such as strokes, amputations and amyotrophic lateral sclerosis may not be able to use their hands yet retaining the ability to speak. Speech-based systems are useful for such users to control the environment and to enhance their education and career opportunities. Nevertheless, speech recognition systems are not widely used as HCI due to the intrinsic sensitivity of such systems to variations in acoustic conditions. The performance of audio speech recognizers degrades when the sound signal strength is low, or in situations with high ambient noise levels. Video recordings of a speaker contain information on the visible movement of the speech articulators such as lips, facial muscles, tongue and teeth. Research where audio and video inputs are combined to recognize large vocabulary, complex speech patterns are being reported in the literature

Matthews et.al. [81] integrated speech cues from many sources and this improves intelligibility, especially when the acoustic signal is degraded. They compared. three methods for parameterizing lip image sequences for recognition sing
hidden Markov models.

Gon|alves et al [82] proposed a segmentation technique for the lips contours together with a set of features based on the extracted contours which is able to perform lip reading with promising results. But its accuracy is very low as for the complete set of experiments the success rate is 35%.

Yau et al. [83] presented a vision-based approach to recognize speech without evaluating the acoustic signals. The proposed technique combines motion features and support vector machines (SVMs) to classify utterances. Potential applications of such a system are, e.g., human computer interface (HCI) for mobility-impaired users, lip-reading mobile phones, in-vehicle systems, and improvement of speech-based computer control in noisy environments.

Saenko et al. [84] presented an approach to detecting and recognizing spoken isolated phrases based solely on visual input. that first employs discriminative detection of visual speech and articulator features, and then performs recognition using a model that accounts for the loose synchronization of the feature streams.

Wang et al. [85] also proposed a new speech recognition method using these visual features and Hidden Markov Model(HMM). Based on global optimization, a new Genetic Algorithm(GA) for training HMM was proposed.

Kubota et al. [86] presented the design and implementation of 3D Auditory Scene Visualizer based on the visual information seeking mantra, i.e., "overview first, zoom and filter, then details on demand". The average recognition rate is 91.47% using improved HMM and 88.96% using the classic training HMM algorithm

Kumar et.al. [87] proposed a time evolution model for AV features and derive an analytical approach to capture the notion of synchronization between them. They proposed this technique to attack the problem of detecting audiovisual (AV) synchronization in video segments containing a speaker in frontal head pose.

Jiang et al. [88] presented a mouth animation construction method based on the DBN models with articulatory features (AF_AVDBN), in which the articulatory features of lips, tongue, glottis/velum can be asynchronous within a maximum asynchrony constraint to describe the speech production process more reasonably. They achieved for AF_ADBN ,AF_AVDBN and S_DBN accuracy of 91.82% ,93.64% and 95.54% respectively.

Massaro et al. [89] designed a device seamlessly worn by the listener, which will perform continuous real-time acoustic analysis of his or her interlocutor's speech. This device would transform several continuous acoustic features of the talker's speech into continuous visual features, which will be simultaneously displayed on the speech reader's eye glasses.
Bregler et al. [90] showed how an SVM based acoustic speaker verification system can be significantly improved in incorporating new visual features that capture the speaker's "Body Language."

### 3.4. Related Work for the Arabic Speech Recognition

Arabic is currently the sixth most widely spoken language in the world. The estimated number of Arabic speakers is 250 million; of which roughly 195 million are first language speakers and 55 million are second language speakers. Arabic is an official language in more than 22 countries. Since it is also the language of religious instruction in Islam, many more speakers have at least a passive knowledge of the language. It is important to realize that what we typically refer to as "Arabic" is not single linguistic variety; rather, it is a collection of different dialects and sociolects, literary form of the language, exemplified by the type of Arabic used in the Quran. Modern Standard Arabic (MSA) is a version of Classical Arabic with a modernized vocabulary. MSA is a formal standard common to all Arabic-speaking countries. It is the language used in the media (newspapers, radio, TV), in official speeches, in courtrooms, and, generally speaking, in any kind of formal communication. However, it is not used for everyday, informal communication, which is typically carried out in one of the local dialects.

Previous work in Arabic ASR has almost exclusively focused on the recognition of Modern Standard Arabic. Some of this work was done as part of the development of dictation applications, such such as the IBM ViaVoice system for Arabic. More recently, information extraction from Broadcast News [91] J. Billa et al. has emerged as an application for Arabic ASR. The current performance of the BBN Arabic Broadcast News recognizer is around 15% word error rate (WER). Johns-Hopkins University Summer Research Workshop Final Report ,2002 developed techniques for making formal Arabic speech data usable for conversational speech recognition, and designed a new statistical models that exploit the available data as well as possible.

Sagheer et al. [92] proposed a visual speech recognition approach based on Hyper Column Model (HCM). The extracted features are modeled by Gaussian distributions through using Hidden Markov Model (HMM). The proposed system, HCM and HMM, can be used for any visual recognition task. They achieved accuracy of about 76.4.

Kirchhoff et al. [93], introduce an Arabic voice recognition system where both training and recognizing process use Romanized characters.] investigate the recognition of dialectal Arabic and study the discrepancies between dialectal and formal Arabic in the speech recognition point of view.

Vergyri et al. [94] investigate the use of morphology-based language model at different stages in a speech recognition system for conversational Arabic; they studied also the automatic diacritizing Arabic text for use in acoustic model training for ASR. In their previous papers Satori et al. [95].

Ananthakrishnan et al. [96] presented generative techniques for recovering vowels and other diacritics that are contextually appropriate, and achieved 70% word level accuracy.
Ramzi A. Haraty and Omar El Ariss [97] concentrated on the Lebanese dialect. The system starts by sampling the speech, which was the process of transforming the sound from analog to digital and then extracts the features by using the Mel-Frequency Cepstral Coefficients (MFCC). The extracted features are then compared with the system's stored model; in this case the stored model chosen was a phoneme-based model.

Ali et al. [98] presented a rule-based technique to generate Arabic phonetic dictionaries for a large vocabulary speech recognition system. The system used classic Arabic pronunciation rules, common pronunciation rules of Modern Standard Arabic, as well as morphologically driven rules.

[99] presented a words Speaker-independent hidden markov models (HMMs)-based speech recognition system was designed using Hidden markov model toolkit (HTK). The database used for both training and testing consists from forty-four Egyptian speakers. Experiments show that the recognition rate using syllables outperformed the rate obtained using monophones, triphones and words by 2.68%, 1.19% and 1.79% respectively.

Ajami et. al. [100] designed a system to recognize an isolated whole-word speech. The Hidden Markov Model Toolkit (HTK) is used to implement the isolated word recognizer with phoneme based HMM models words. Speaker-independent hidden markov models (HMMs)-based speech recognition system was designed using Hidden markov model toolkit (HTK). The database used for both training and testing consists from forty-four Egyptian speakers. Experiments show that the recognition rate using syllables outperformed the rate obtained using monophones, triphones and words by 2.68%, 1.19% and 1.79% respectively. The selected monophone-based recognition rate is 90.75%. The selected triphone based recognition rate is 92.24%. The selected syllable-based recognition rate is 93.43%. The selected word-based recognition rate is 91.64% using 5-states of HMM-based but at 13-states of HMM-based, the recognition rate is 97.01%.

By the next section, the proposed framework to handle the communication difficulties between the normal and the deaf persons will be discussed.

## 4. THE PROPOSED DEAF-NORMAL-DEAF FRAMEWORK

### 4.1. Research and Challenges in Automatic Sign Language Recognition

In the following points the building up steps for a large vocabulary Arabic sign language recognition system.

• *Environment Conditions and Feature Extraction.*

The main difficulty for a sign language (SL) recognition is extracting the sign features from the environmental conditions such as difficult backgrounds and lightning conditions. Also , as discussed in the previous sections, many current systems are used to recognize the SL but most of them depend on data gloves, colored gloves, or extracting the sign from a simple environment.

• *Signs Modeling Facilities*

To build up a continuous sign language recognition system, different modules are required to corporate in order to generate sign recognition. These modules are discussed in the next subsection.

Once the different modules are built; the following application scenarios would be possible:
- E-learning of sign language
- Automatic transcription of video e-mails, video documents, or video-SMS
- Video subtitling and annotation

The novel features of such systems provide new ways for solving industrial problems. The technological breakthrough of SignsWorld will clearly impact on other applications fields:
- Improving human-machine communication by gesture: vision-based systems are opening new paths and applications for human-machine communication by gesture, e.g. Play Station's
- Medical field: Communication techniques may be being used to improve the communication between the medical staff, and any electronic equipment.
Another application in this sector is related to web- or video-based e-Care / e-Health
Treatments or an auto-rehabilitation system which makes the guidance process to a patient during the rehabilitation exercises easier.
- Surveillance sector: person detection and recognition of body parts or dangerous objects, and their tracking within video sequences or in the context of quality control and inspection in manufacturing sectors.

### 4.2. SignsWorld Specifications

The SignsWorld is the proposed project that will navigate in the signs and deaf world. The overall goal of the SignsWorld project is to develop a new vision-based technology for recognizing and translating continuous Arabic sign language ArSL (i.e. provide Video-to-Text technologies), in order to provide new e-Services to the deaf community and to improve their communication with the hearing people. The specifications of SignsWorld are presented below.

*1. Multimodal system.* SignsWorld seeks to explicitly exploit the complementarities and redundancies between the different communication channels (two hands, face, head, upper body), especially in terms of boundary detection.

International Journal of Computer Science & Information Technology (IJCSIT) Vol 4, No 1, Feb 2012

*2. Naturality.* The signer will speak without wearing data gloves, colored gloves or other types of sensors or markers.

*3. Vocabulary size (for ArSL)* around 3.300 words.

*4. Signer-independency*. The system will be similar to robust automatic speech recognition systems. Signer independence also implies pronunciation, language modelling adaptation and the usage of speaker adaptation techniques.

*5. Translation feasibility*. The system won't only identify the isolated signs but also the continuous signs.

*6 Self-adaptation to changing the external conditions*. Additional research will be made to allow the system to work independently of the background colors and the signers' clothes and brightness, in order to enable robust tracking and speed measurements of the targeted body parts.

### 4.3. The SignsWorld Project Architecture

Fig.3. shows the deaf-to-normal communication system architecture which is divided into three phases. The first phase is the signs and gestures preprocessing and training which prepare the input signs and gestures signal, extract their features, and make training for them with taking into consideration whether the signal is from a far distance or from a different angle (homo-graphical training). The second one is the recognition phase that includes the recognition of the sign/gesture type (facial expression, finger spelling, sign language or an epenthesis) with the signer independency. Then the third part is the user interface outputting phase which helps the normal person to understand the deaf person through converting the sign into text or/and speech.

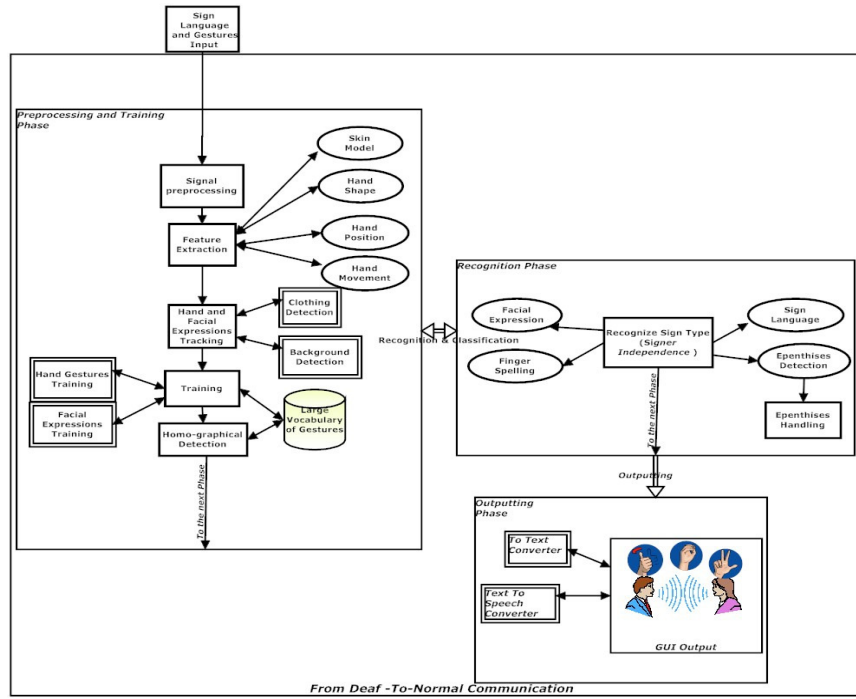

Figure 3 The deaf-to-normal communication system architecture

Figure 4. shows the normal-to-deaf communication system architecture which is divided into four phases. The first phase is the audio/visual signal preprocessing and training which prepare the input signal, extract their features, track the video signal, segment the video or audio signal individually and finally extract the noise signals form them. The second phase is the training phase in which the system is trained for the Arabic audio and visual patterns, emotions are trained, and also the grammar and model patterns are saved. The third one is the recognition phase that includes the recognition of the Arabic word (speaker independency), and the emotion type (sadness, anger, happiness, neutral, mean, ..etc). Then the final phase is the user interface outputting phase which helps the deaf person to understand the normal person through converting the audio/visual speech into sign language, his native language.

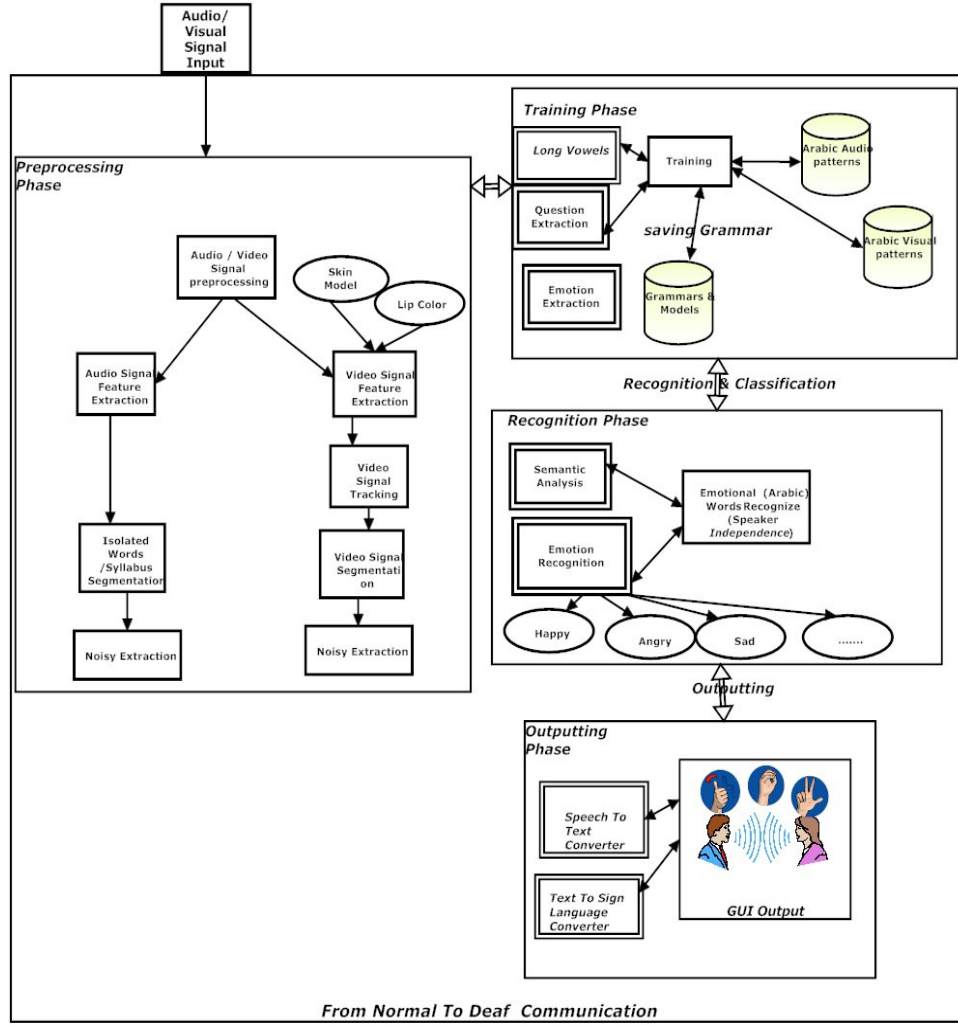

Figure 4.  The normal-to-deaf communication system architecture

## 5. CONCLUSIONS AND FUTURE WORK

This paper presented the most popular techniques and technologies used for enabling the deaf people to communicate with the normal people easily and vice versa. By studying of sign language recognition ,and the Arabic sign language recognition techniques in specific, we found that there is no complete system that can interpret a large vocabulary of Arabic manual and non-manual signs. Existing techniques mainly focus on the detection of only static manual postures (without NMS) or those dynamic gestures which do not involve any motion in the third dimension. Also studying the speech and visually speech recognition techniques told us that there is no complete Arabic visually emotionally speech recognition system that includes the Arabic grammar models. Consequently, we proposed a general framework for the SignsWorld system that is proposed to facilitate the communication between the Arabic deaf and normal people.